# LeCoT: revisiting network architecture for two-view correspondence pruning

Luanyuan Dai, Xiaoyu Du & Jinhui Tang*

*School of Computer Science and Engineering, Nanjing University of Science and Technology, Nanjing 210094, China*

**Abstract** Two-view correspondence pruning aims to accurately remove incorrect correspondences (outliers) from initial ones and is widely applied to various computer vision tasks. Current popular strategies adopt multilayer perceptron (MLP) as the backbone, supplemented by additional modules to enhance the network ability to handle context information, which is a known limitation of MLPs. In contrast, we introduce a novel perspective for capturing correspondence context information without extra design modules. To this end, we design a two-view correspondence pruning network called LeCoT, which can naturally leverage global context information at different stages. Specifically, the core design of LeCoT is the Spatial-Channel Fusion Transformer block, a newly proposed component that efficiently utilizes both spatial and channel global context information among sparse correspondences. In addition, we integrate the proposed prediction block that utilizes correspondence features from intermediate stages to generate a probability set, which acts as guiding information for subsequent learning phases, allowing the network to more effectively capture robust global context information. Notably, this prediction block progressively refines the probability set, thereby mitigating the issue of information loss that is common in the traditional one. Extensive experiments prove that the proposed LeCoT outperforms state-of-the-art methods in correspondence pruning, relative pose estimation, homography estimation, visual localization, and 3D reconstruction tasks. The code is provided in https://github.com/Dailuanyuan2024/LeCoT-Revisiting-Network-Architecture-for-Two-View-Correspondence-Pruning.

**Keywords** correspondence pruning, transformer, spatial and channel information, context information, relative pose estimation

## 1 Introduction

Finding good-quality two-view correspondences is an elementary task in computer vision, which strives to build sparse correspondences between images from two different views and estimate geometry relationship. It serves as the precondition for various challenging computer vision tasks, *e.g.*, visual localization [1], Simultaneous Location and Mapping (SLAM) [2], point cloud registration [3], etc.

The most common geometry matching pipeline includes two stages: feature extraction and matching, and correspondence pruning. In the initial stage, researchers expend considerable effort on local feature descriptors (such as SIFT [4] and SuperPoint [5]), and subsequently construct a putative correspondence set based on their similarities. Unfortunately, there are massive outliers because of rotations, viewpoint changes and illumination changes in image pairs. To mitigate the adverse effects of outliers, it is customary to eliminate them and choose a dependable subset predominantly comprising correct correspondences (*i.e.*, inliers) to robustly estimate the relative pose. The process, named correspondence pruning, aims to remove outliers to precisely gain correspondences and geometry relationships.

Traditional correspondence pruning methods such as Random Sample Consensus (RANSAC) [6], Vector Field Consistency (VFC) [7], Locality Preserving Matching (LPM) [8] and its variants [9–11] typically rely on handcrafted features and heuristic rules to identify and remove outliers in specific scenes. However, in real-world scenarios, there are complex scene variations and considerable outliers, which often render them ineffective. In contrast, deep learning-based techniques possess stronger learning and generalization capabilities, enabling them to learn abstract features and high-level representations from vast amounts of data. Therefore, researchers have turned their attention to deep learning-based techniques. In CNe [12], correspondence pruning is treated as a binary classification problem under the multilayer perceptron (MLP) framework. Unfortunately, MLP processes each sparse correspondence independently, as shown in Figure 1(a). The intrinsic individual feature extraction capability of MLP poses a significant problem: the lack of context information, which is crucial for handling two-view correspondences and geometry learning [13]. Therefore, current popular approaches [14–21] design additional modules, inserted into the MLP backbone, to capture context information, as depicted in Figure 1(a). Despite the promising performance has been attained with this framework, drawbacks still persist. Apparently, these challenges associated with these supplementary modules have impeded the acquisition and consolidation of context information, consequently restricting the performance of this framework (see in Figure 1(a)).

---

* Corresponding author (email: jinhuitang@njust.edu.cn)

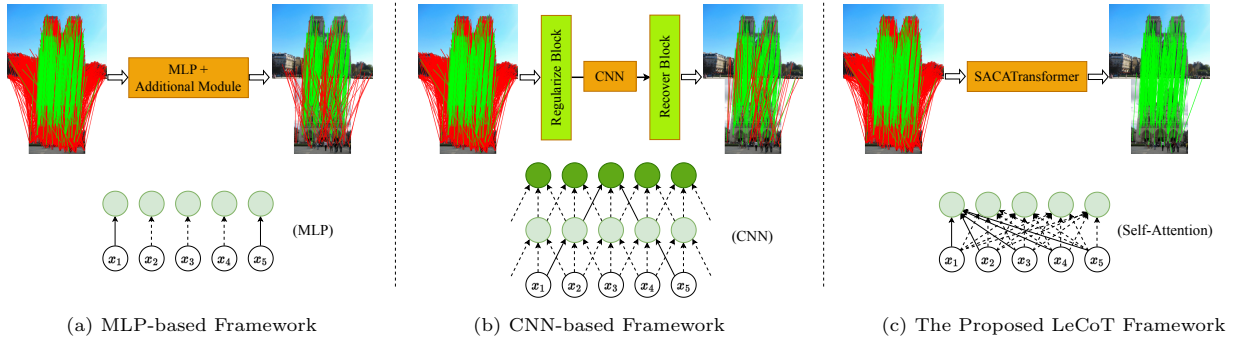

**Figure 1** Comparative illustrations of correspondence pruning for three distinct frameworks: (a) MLP-based, (b) CNN-based, and (c) the proposed LeCoT, where the corresponding diagrams below are MLP, CNN, and self-attention (a crucial component of the Transformer architecture), respectively. The arrows represent the information transmission. The solid lines in (a) indicate that MLP processes each correspondence separately. The solid lines in (b) indicate the information transmission of a CNN with a convolutional kernel size of $3 \times 3$ and stride 1. As the number of CNN layers increases, the receptive field gradually expands. The solid lines in (c) illustrate the information propagation of self-attention, where attention distributions are computed at each layer to determine the correlation between any two correspondences, thus aggregating global context information.

Moreover, ConvMatch [13] advocates for the adoption of Convolutional Neural Network (CNN) as the backbone to tackle this challenge, due to its ability to gradually integrate local and global information. While this approach partly alleviates the issue of MLP neglecting relationships among correspondences, some problems still endure. Theoretically, increasing the number of CNN layers can capture global context information from local context information. However, when considering the YFCC100M dataset [22] for relative pose estimation in ConvMatch [13], which involves $N = 2000$ correspondences and utilizes the specified CNN architecture (with a $3 \times 3$ convolutional kernel, a stride of 1, and no padding) as depicted in Figure 1(b), achieving comprehensive global context information would require approximately 1000 layers. This is obviously unrealistic for this task. Fortunately, ConvMatch [13] converts unordered correspondences into a $16 \times 16$ vector field (**data compression**, as shown in 1(b)), but this still requires 8 CNN layers to capture global context information. However, the study only uses 6 CNN layers, making it **difficult to capture global context information adequately**. Furthermore, the forced conversion of unordered data into ordered ones (**data compression**) can result in **information loss and, disrupt the inherent structure and distribution of the data** [23–25]. Additionally, the local perception mechanism in CNN may lead to **a gradual loss of information propagation within the network**, particularly concerning correlated information from distant positions [26, 27]. In this scenario, one might wonder how to further address this issue: *whether to continue refining the context-capturing module to compensate for the inherently context-agnostic MLP, deepen the CNN-based network, or replace it directly with a network inherently possessing stronger context perception capabilities?*

Transformer inherently possesses the outstanding capability to effectively capture global context information, as evidenced by their remarkable performance in computer vision tasks such as image classification [28] and semantic segmentation [29, 30]. Hence, we opt for Transformer over MLP or CNN to extract deep features for sparse correspondences, which can naturally address the issue of insufficient global context information among correspondences without additional module designs and data compression, see in Figure 1(c). Transformer is originally designed to process ordered data, primarily in Natural Language processing (NLP). However, sparse correspondences are unordered, so we choose Transformer without positional encoding. Following the idea, we construct a novel and straightforward two-view correspondence pruning framework, named LeCoT, which can naturally harness global context information from various stages without extra modules and data compression.

In conclusion, our primary contributions can be outlined as follows:

• From a new perspective, we propose a simple and effective two-view correspondence pruning framework LeCoT, without additional module designs and data compression, which can naturally utilize global context information at various stages.

• We introduce a novel Spatial-Channel Fusion Transformer block as the backbone to simultaneously leverage spatial and channel global context information among sparse correspondences. To our knowledge, this is the first instance of utilizing the complete Transformer structure as the backbone to tackle the correspondence pruning problem.

• We use the proposed prediction block to progressively obtain the probability set from correspondence features learned at the intermediate stage, which can serve as guiding information to find robust global context information.

• The proposed LeCoT consistently surpasses current state-of-the-art networks in correspondence pruning, relative pose estimation, homography estimation, visual localization, and 3D reconstruction.

The remaining of this paper is organized as follows: In Section 2, we review correspondence pruning and Transformer related works. In Section 3, we present the overall framework and details of the proposed LeCoT. In Section 4, we show the experimental results. In Section 5, we make a conclusion.

## 2 Related work

### 2.1 Learning-based correspondence pruning networks

PointNet [31] pioneers learning-based approaches by employing an MLP-based backbone to classify and segment irregular point clouds. CNe [12] draws inspiration from the PointNet-like architecture, which is the first one to treat correspondence pruning as a binary classification problem with the MLP backbone. But the MLP backbone fails to fully capture context information from sparse correspondences, so various additional modules and strategies have been proposed.

Some networks [15, 16, 19, 32, 33] draw lessons from the attention machine [34] to aggregate information locally or globally. Specifically, ACNe [15] introduces a novel context normalization mechanism called Attentive Context Normalization (ACN), which captures context information by applying attention weighting to local regions, thereby enhancing the model comprehension of input data. LAGA [16] captures context information locally and globally by the attention mechanism to enhance the network ability. MS$^2$DGNet [19] leverages attention to cluster information within the constructed sparse semantic dynamic graph to enhance the network performance. Other networks [14, 17–19, 35–37] are influenced by GNNs to construct graphs in local or global levels. In particular, OANet [14] enhances the ability to handle spatial geometric transformations and ordering information between views effectively by incorporating a clustering layer into the intermediate of CNe [12], where the clustering layer is implemented based on GNNs. LMCNet [17] employs GNNs to achieve motion coherence and extends it as a differentiable layer in the network. CLNet [18] and U-Match [38] sequentially integrate local and global contexts by GNNs, progressively eliminating outliers. Based on CLNet [18], BCLNet [36] addresses the issue of ignoring interactions between different contexts. GCT [37] combines a Transformer and GNNs on an MLP-based backbone to provide additional information. TrGa [39] addresses the issue of information confusion in GNNs and adopts Transformers with stronger feature extraction capabilities as the backbone in correspondence pruning. In addition, LFC [40] and DeMatch [41] introduce a Siamese network and a diffusion model into the correspondence pruning process, respectively.

While the mentioned networks have advanced in the network structure and shown good performance, they all rely on the MLP-based backbone, which struggles to address the challenge of being context-agnostic. Hence, ConvMatch [13] proposes converting unordered correspondences into ordered vector fields and employs CNN as the backbone to tackle this issue. While it has made some progress, the conversion of unordered correspondences to ordered vector fields may result in information loss and disrupt the inherent structure and distribution of the data. Furthermore, the local perception mechanism in the CNN may lead to gradual loss of information propagation, especially concerning distant information, making it difficult to capture global context information. Therefore, we construct a simple yet effective two-view correspondence pruning network LeCoT, from a fresh perspective, without additional modules and data compression, which can naturally leverage global context information.

## 2.2 Transformers in vision related tasks

Vision Transformer (ViT) [28] first bridges the gap between Natural Language Processing (NLP) and Computer Vision (CV) by employing the standard Transformer [34] to handle full-sized images, achieving state-of-the-art results in the ImageNet [42] classification task. Inspired by the success of ViT, numerous diverse visual Transformers have been proposed for various computer vision tasks, yielding remarkable results. Compared to ViT [28], Image Transformer [43] boasts a more streamlined design and introduces several improvements, including a local self-attention mechanism and dynamic feature map resizing. TransUNet [44] is the first medical image segmentation framework that combines a Transformer with the U-Net, fully leveraging the strengths of the Transformer in capturing global information and the U-Net in extracting local features, thereby achieving powerful image segmentation performance. SSFormers [45] is proposed for few-shot learning, which can identify task-relevant features and suppress task-irrelevant features. Segmenter [29] is the first one to successfully apply the Transformer architecture to the semantic segmentation task, achieving notable success in pixel-level image segmentation tasks. TransCrowd [46] redefines the weakly-supervised crowd counting problem from the perspective of sequence-to-count based on Transformer. Pyramid Vision Transformer (PVT) [47] introduces a pyramid-style attention mechanism, enabling the model to better handle image features at different scales. T2T-ViT [48] introduces a novel approach allowing training vision Transformer models from scratch on the ImageNet dataset, leveraging the strengths of both Transformer and CNN architectures, achieving performance comparable to pre-trained Vision Transformers without using pre-trained weights. DETR [49] is an end-to-end object detection network without redundancy boxes, which greatly optimizes and promotes the development of object detection. Swin Transformer [50] uses shifted windows to obtain multiple scale information to deal with different size of objects in each image. Inspired by BERT [51], MAE [52] masks a high proportion of each input image, so that it can achieve self-supervised learning. ViLT [53] uses Transformers well in the vision-and-language task, which greatly reduces network running cost and has comparable performance. To the best of our knowledge, the proposed LeCoT is the first two-view correspondence pruning network with the Transformer-based backbone, building upon the highly successful ViT [28]. Moreover, Transformers have also shown strong cross-modal modeling capability in tasks such as video-text retrieval [54,55], video-audio alignment [56], large-scale video understanding [57], robotic affordance reasoning [58], and vision–tactile–language–action learning [59].

## 3 Methodology

### 3.1 Problem formulation

Given a pair of matching images $(I, I')$, we first use local features (SIFT [4], SuperPoint [5], etc.) to detect keypoints and obtain descriptors. After that, we build an initial correspondence set $C$ by an NN matcher:

$$C = \{c_1; c_2; ...; c_N\} \in \mathbb{R}^{N \times 4}, \tag{1}$$

where $c_i = (x_i, y_i, u_i, v_i)$ is defined as a correspondence between two keypoints $(x_i, y_i)$ and $(u_i, v_i)$ at the corresponding positions of the image pair, and both of them are normalized under camera intrinsics. But unfortunately, as mentioned earlier, the initial correspondence set usually contains a large portion of outliers. Therefore, our job is to eliminate outliers while preserving inliers as much as possible.

In this paper, we use the proposed LeCoT to achieve this goal. As shown in Figure 2, we introduce an improved verification framework inspired by [6, 18], which has two pruning modules. Our goal is to gain the final probability set $P = \{p_1; p_2; ...; p_N\}$ with $p_i \in [0, 1)$. Specifically, a guiding information probability set $P_s$ is proposed to guide and adjust the training process. After that, we obtain the pruned guiding information probability set $\hat{P}_s$, the first estimated probability set $\hat{P}_1$ and the first subset $\hat{C}_1$ with a higher inlier rate after the first pruning module. Next, $\hat{P}_s$, $\hat{P}_1$ and $\hat{C}_1$ are used to produce $\hat{P}_2$ and $\hat{C}_2$ in the second pruning module. Then, a weighted eight-point algorithm [12] (Model Estimation in Figure 2) is used to calculate an essential matrix $\hat{E}$. Ultimately, we use a verification operation to count the final probability set $P$. The above process can be written as:

$$\begin{aligned}(\hat{P}_s, \hat{P}_1, \hat{C}_1) = f_{1\phi}(C), \quad (\hat{P}_2, \hat{C}_2) = f_{2\psi}(\hat{P}_s, \hat{P}_1, \hat{C}_1),\\ \hat{E} = g(\hat{P}_2, \hat{C}_2), \quad P = Ver(\hat{E}, C),\end{aligned} \tag{2}$$

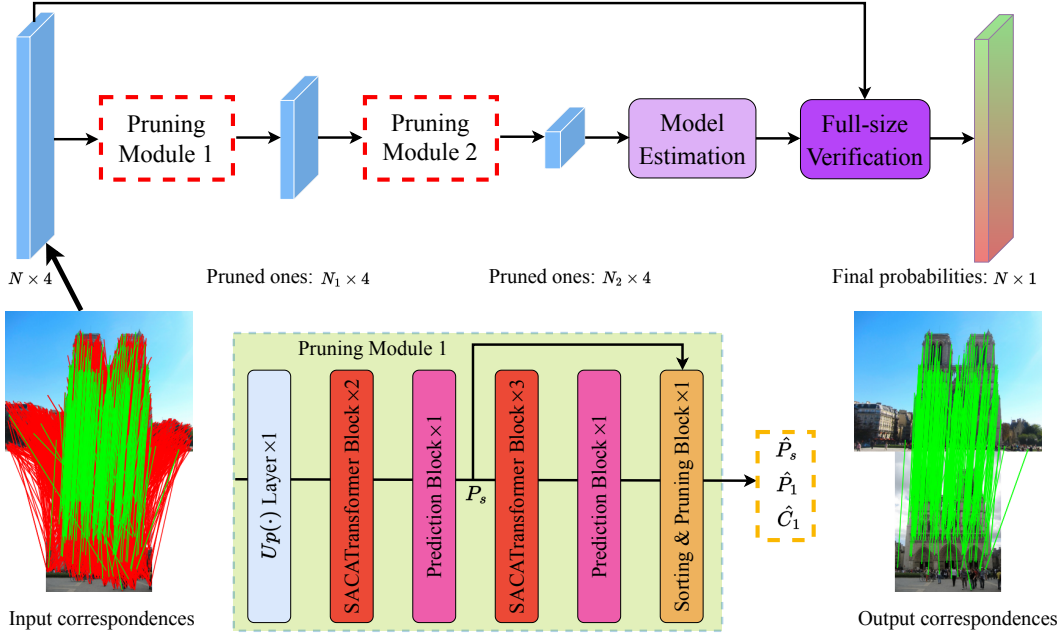

**Figure 2** The pipeline of the proposed LeCoT. The input is an initial correspondence set $C \in \mathbb{R}^{N \times 4}$ and the output is the final probability set $P \in \mathbb{R}^{N \times 1}$ through an iterative pruning strategy. This process aims to distill more reliable candidates to estimate the parametric model. Each pruning module comprises an $Up(\cdot)$ layer for dimensionality transformation, two proposed prediction blocks, and five proposed spatial-channel fusion transformer (SACATransformer) blocks, and a sorting & pruning block. $P_s$, $\hat{P}_s$, $\hat{P}_1$ and $\hat{C}_1$ are the guiding information probability set, the pruned guiding information probability set, the first estimated probability set and the first subset with a higher inlier rate after the first pruning module, respectively.

where $f_{1\phi}(\cdot)$ and $\hat{P}_1$ are the first pruning module with learnable parameter $\phi$ and the first estimated probability set, respectively; $f_{2\psi}(\cdot,\cdot)$ and $\hat{P}_2$ are the second pruning module's; $g(\cdot,\cdot)$ is denoted as the weighted eight-point algorithm; $Ver(\cdot,\cdot)$ is the full-size verification operation, in which correspondences that satisfy the estimated model $\hat{E}$ are labeled as inliers, otherwise labeled as outliers.

### 3.2 Main structure of LeCoT

The overall LeCoT architecture is simple and shown in Figure 2, which includes three main components: two pruning modules, a model estimation module, and a full-size verification module. The model estimation module utilizes the last pruned reliable correspondence subset $\hat{C}_2 \in \mathbb{R}^{N_2 \times 4}$ to estimate the essential matrix, which is then passed to the full-size verification module to classify all correspondences. Particularly, each pruning module is made up of an $Up(\cdot)$ layer, five SACATransformer blocks, two prediction blocks and a Sorting&Pruning block. Firstly, due to the low-dimensionality of the correspondence set hindering the extraction of deep features, as a routine procedure, we convert it into a high-dimensional correspondence feature set $F \in \mathbb{R}^{N \times d}$ as the input correspondence feature set.

$$F = Up(C), \tag{3}$$

where the $Up(\cdot)$ layer is utilized to map the correspondence set from a low-dimensional space into a high-dimensional space. We opt for a simple Multi-Layer Perceptron (MLP) to fulfill this task.

Then, we utilize SACATransformer blocks to capture spatial and channel global context information for the unordered correspondence features, rather than context-agnostic MLP. In this process, we use the proposed prediction block (see Section 3.5) to obtain the guiding information $P_s$, which is then passed to the Sorting&Pruning block to obtain the pruned guiding information $\hat{P}_s$, which is one of the inputs of the second pruning module. By integrating the aforementioned modules, we establish a progressive correspondence pruning network.

### 3.3 Spatial-channel fusion transformer

A vanilla Transformer block consists of two PreNorms (PNs), one Multi-Head Self-Attention (MSA) and a FeedForward (FF), as depicted in Figure 3(a). We choose the layer normalization as the PreNorm

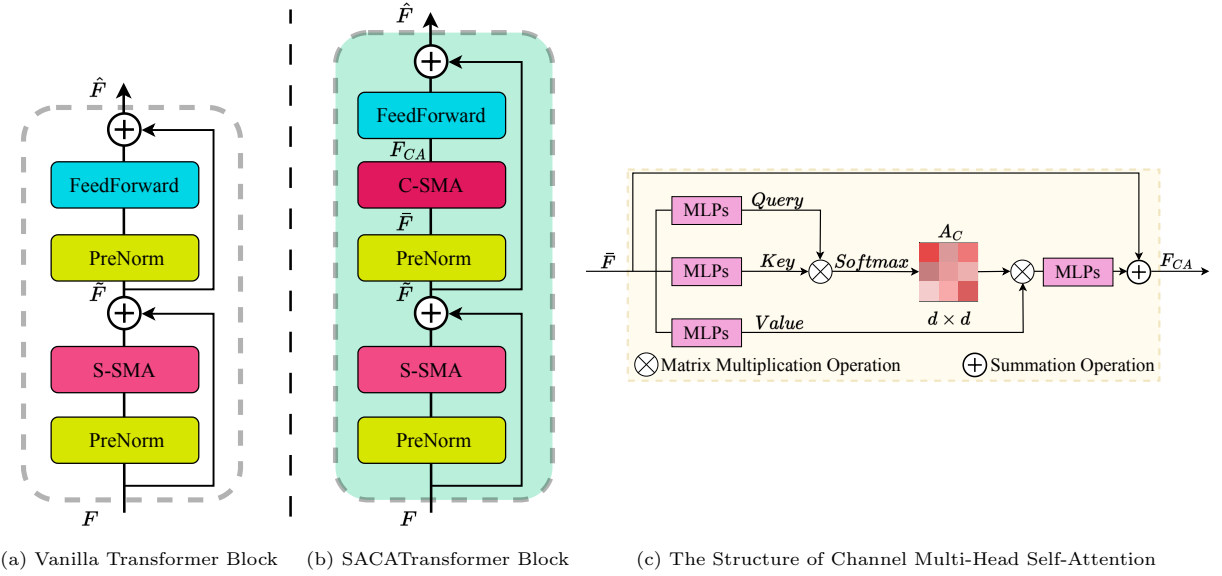

(a) Vanilla Transformer Block  (b) SACATransformer Block  (c) The Structure of Channel Multi-Head Self-Attention

**Figure 3** Structures of the vanilla Transformer block and the SACATransformer block. (a) is the vanilla transformer block obtained from ViT. (b) is the proposed SACATransformer block, which can capture global context information from spatial and channel dimensions. (c) is Channel Multi-Head Self-Attention (C-MSA).

(PN), because it can better handle variable length data. Followed by ViT [28], a FeedForward (FF) is made up of two linear layers and a GELU layer. Notably, the multi-head self-attention mechanism in the vanilla Transformer block typically operates on the spatial dimension of the input sequence. Therefore, we refer to it as Spatial Multi-Head Self-Attention, denoted as S-MSA, as illustrated in Figure 3. Taking the correspondence feature set $F$ as an example, which is successively put into a PreNorm and a S-MSA followed by a residual structure. After that, the output $\tilde{F}$ is passed through a PreNorm and a FeedForward followed by a residual structure to obtain $\hat{F}$. The above operations can be written as:

$$\tilde{F} = S\text{-}MSA\left(PN\left(F\right)\right) + F, \tag{4}$$

$$\hat{F} = FF\left(PN\left(\tilde{F}\right)\right) + \tilde{F}. \tag{5}$$

However, we find that using the vanilla Transformer is not optimal because it lacks consideration of the global context information about correspondence features in the channel dimension. Therefore, we incorporate the proposed Channel Multi-Head Self-Attention, denoted as C-MSA, between the PreNorm and the FeedForward of the vanilla Transformer, as shown in Figure 3(b). The reconfigured block is termed as Spatial-Channel Fusion Transformer, abbreviated as SACATransformer. Thus, Eq. (5) can be rewritten as:

$$\hat{F} = FF\left(C\text{-}MSA\left(PN\left(\tilde{F}\right)\right)\right) + \tilde{F}. \tag{6}$$

We will provide a detailed explanation of Channel Multi-Head Self-Attention (C-MSA) in Section 3.4

### 3.4 Channel multi-head self-attention

We modify the standard multi-head self-attention mechanism [34] by computing $Query$, $Key$ and $Value$ along the channel dimension. As illustrated in Figure 3(c), we first process the input features using simple MLPs to obtain $Query$, $Key$ and $Value$. Then, we compute the similarity matrix $A_C \in \mathbb{R}^{d \times d}$ by multiplying $Query$ and $Key$, followed by a matrix multiplication operation between the similarity matrix $A_C$ and $Value$. Finally, the resulting output is passed through MLPs to further learn abstract features within the input, followed by element-wise summation. The specific steps are as follows:

$$Query, Key, Value = MLPs(\tilde{F}), \tag{7}$$

$$A_C = Softmax\left(Query \times Key^T\right), \tag{8}$$

$$F_{CA} = \sum\left(A_C \times Value\right) + \tilde{F}, \tag{9}$$

where $\times$ and $Softmax(\cdot)$ present a matrix multiplication operation and a softmax operation, respectively; $\sum$ represents the element-wise summation operation.

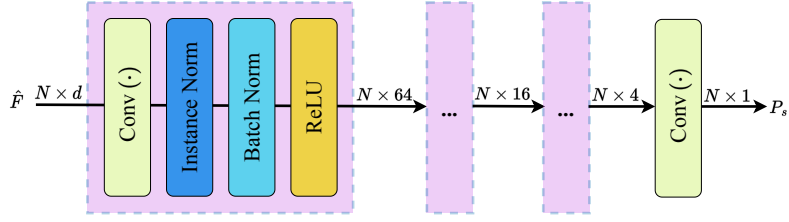

**Figure 4** The structure of the prediction block. The input is the correspondence feature $\hat{F}$ obtained by the first two SACATransformer blocks and the output is the guiding information $P_s$.

### 3.5 Prediction block

In Figure 2, we utilize the proposed prediction block to obtain the probability set as the guiding information, which can be written as:

$$P_s = Pre\left(\hat{F}\right), \quad (10)$$

where $Pre(\cdot)$ represents the proposed prediction block, which includes Conv($\cdot$), Instance Norm, Batch Norm and ReLU functions.

A common practice, such as in OA-Net++ [14], is to directly use a projection function to map the channel dimension from $d$ to 1 as a basis for judging the correctness of correspondences. However, this approach may result in significant information loss and may not be sufficient to fully represent the input features [23]. Furthermore, as observed from Table 10, the network employing the proposed prediction block yields superior performance compared to directly using a simple projection network, like Conv($\cdot$). Therefore, we opt for gradually reducing the dimensionality, as shown in Figure 4, which can mitigate information loss.

### 3.6 Loss function

Following previous works [14,18], we select a hybrid loss function to iteratively train the proposed LeCoT:

$$L_{oss} = L_e(\hat{E}, E) + \beta L_c, \quad (11)$$

where $L_e$ is a geometric loss between an estimated model $\hat{E}$ and the true one $E$; $\beta$ is a weighting factor to balance these two losses; $L_c$ is a binary classification loss with a proposed adaptive temperature [18] and can be defined as:

$$L_c = \sum_{i=1}^{I} \left( H(\sigma(\tau_i \odot \hat{P}_s), y) + H(\sigma(\tau_i \odot \hat{P}_i), y) \right), \quad (12)$$

where $I$ is the iteration number; $H(\cdot, \cdot)$ denotes a binary cross entropy function; $\tau_i$ is an adaptive temperature vector to mitigate the influence of label ambiguity; $\hat{P}_s$ and $\hat{P}_i$ are the pruned guiding information and the $i$-th pruning module probability set, respectively; $y$ is the binary ground truth label set; $\odot$ and $\sigma(\cdot)$ represent a Hadamard product and a sigmoid function, respectively. $L_e$ is a geometric loss, which can be written as:

$$L_e(E, \hat{E}) = \frac{(p'^T \hat{E} p)^2}{\|Ep\|_{[1]}^2 + \|Ep\|_{[2]}^2 + \|E^T p'\|_{[1]}^2 + \|E^T p'\|_{[2]}^2}, \quad (13)$$

where $E$ and $\hat{E}$ are the ground truth and estimated essential matrices, respectively; $p$ and $p'$ are virtual correspondence coordinates obtained by the ground truth $E$.

### 3.7 Implementation details

Network input is an $N \times 4$ correspondence set, where $N$ is up to 2000. There are two pruning modules in total, with a pruning rate of 0.5, resulting in a final set of $N/4$ reliable correspondences. Channel dimension $d$ is 128. SACATransformer block number $L$ is 5, where 4-head attention is used. Batchsize is 32, and $\beta$ in Eq. 11 is 0.5. We choose Adam optimizer with a learning rate of $10^{-3}$. And a warmup strategy is used to train LeCoT. Clearly, a linearly growing rate is used for the first $10k$ iterations, after that the learning rate begins to decrease and reduce for every $20k$ iterations with a factor of 0.4. Experiments are carried on NVIDIA GTX 3090 GPUs.

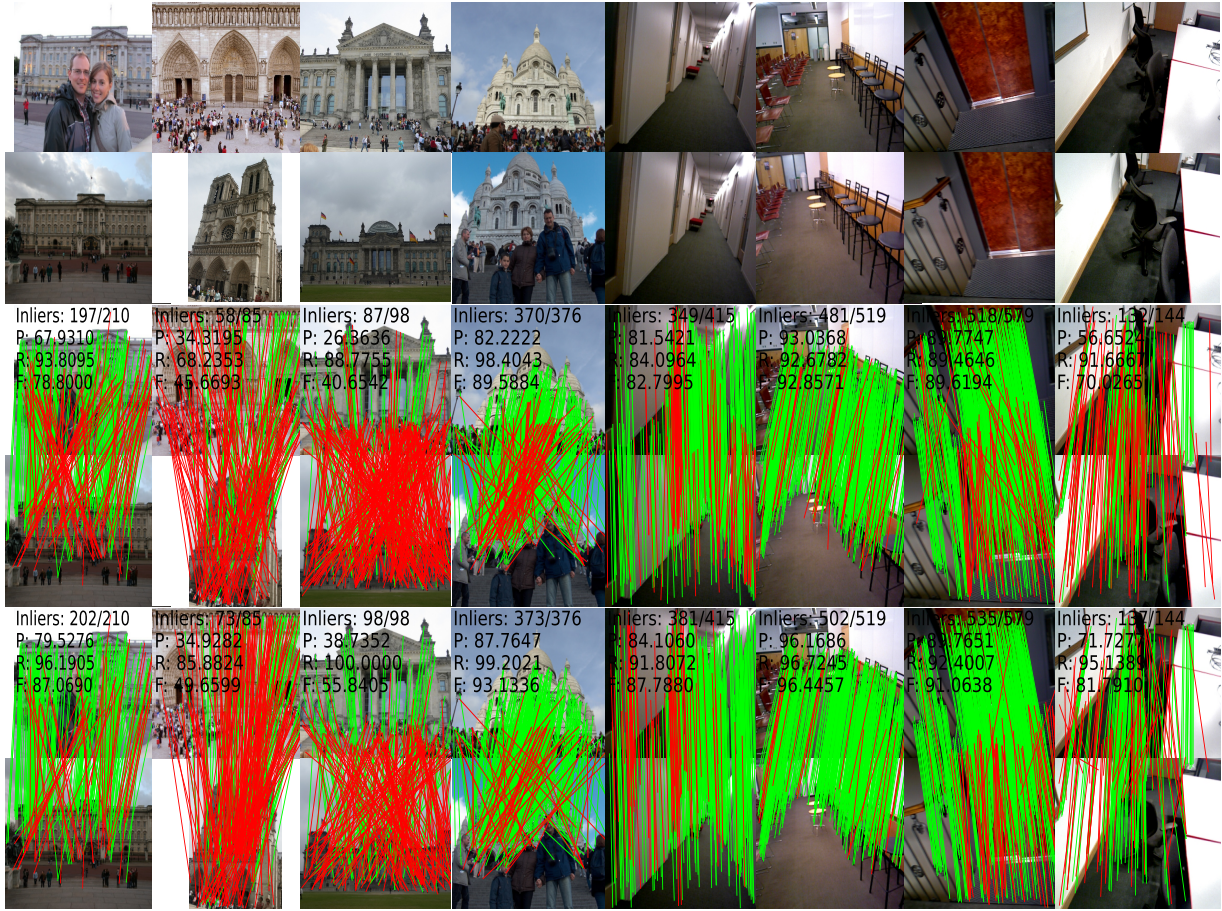

**Figure 5** Partial typical visualization results on YFCC100M and SUN3D datasets with SIFT. From top to bottom: input image pairs, results of MS²DG-Net and the proposed LeCoT. Green and red lines are inliers and outliers, respectively. The top left corner of each image displays $Inliers$, $P$, $R$, and $F$.

## 4 Experiments

### 4.1 Correspondence pruning

Correspondence pruning is the process of identifying and eliminating outliers from initial correspondences to ensure accurate ones, which can enhance the accuracy and robustness of matching between the two views.

**Datasets.** We opt for Yahoo's YFCC100M [22] and SUN3D [60] datasets as outdoor and indoor scenes, respectively. In the outdoor dataset, 68 sequences serve as training data, while the remaining 4 sequences are allocated for testing. For the indoor dataset, 239 sequences are chosen for training, leaving 15 sequences for network evaluation.

**Evaluation protocols.** $Precision$ ($P$), $Recall$ ($R$) and $F$-$score$ ($F$) are selected as metrics in correspondence pruning. $Precision$ ($P$) measures the accuracy of correct matches by dividing the number of correct matches by the total number of samples classified as positive. $Recall$ ($R$) quantifies the completeness of correct matches by assessing the ratio of the number of correct matches to the total number of samples actually belonging to the positive class. $F$-$score$ ($F$) is the harmonic mean of $Precision$ and $Recall$, providing a balanced assessment of a network performance.

**Baselines.** Ten learning-based networks (CNe [12], OA-Net++ [14], ACNe [15], LMC-Net [17], CL-

**Table 1** Quantitative comparative results of correspondence pruning on YFCC100M and SUN3D with SIFT. And the best result in each column is in bold.

| Method | References | Size (MB) | YFCC100M (%) | | | SUN3D (%) | | |
|---|---|---|---|---|---|---|---|---|
| | | | P | R | F | P | R | F |
| CNe [12] | CVPR2018 | **0.39** | 52.84 | 85.68 | 65.37 | 46.11 | 83.92 | 59.37 |
| OA-Net++ [14] | ICCV2019 | 2.47 | 55.78 | 85.93 | 67.65 | 46.15 | 84.36 | 59.66 |
| ACNe [15] | CVPR2020 | 0.41 | 55.62 | 85.47 | 67.39 | 46.16 | 84.01 | 59.58 |
| LMC-Net [17] | CVPR2021 | 0.93 | 60.12 | 87.50 | 71.27 | 47.23 | 84.60 | 60.62 |
| CL-Net [18] | ICCV2021 | 1.27 | 74.89 | 76.79 | 75.83 | 59.97 | 68.35 | 63.89 |
| MS$^2$DG-Net [19] | CVPR2022 | 2.61 | 59.11 | 88.40 | 70.85 | 46.95 | 84.55 | 60.37 |
| ConvMatch [13] | AAAI2023 | 7.49 | 60.03 | 89.19 | 71.76 | 48.01 | **85.19** | 61.41 |
| LFC [40] | ACM MM 2023 | 1.73 | 60.84 | 88.66 | 72.16 | 48.14 | 85.09 | 61.49 |
| GCT [37] | AAAI 2024 | 4.09 | 77.00 | 79.02 | 78.00 | 61.12 | 69.34 | 64.97 |
| DeMatch [41] | CVPR 2024 | 5.86 | 60.48 | **89.79** | 72.28 | 48.25 | 84.72 | 61.48 |
| LeCoT (Ours) | - | 3.74 | **77.70** | 80.43 | **79.04** | 61.00 | 68.77 | **64.65** |

**Table 2** Evaluation on YFCC100M and SUN3D with SIFT for relative pose estimation. mAP5° and mAP20° are reported, and the best result in each column is in bold.

| Method | References | Size (MB) | YFCC100M (%) | | SUN3D (%) | |
|---|---|---|---|---|---|---|
| | | | mAP5° | mAP20° | mAP5° | mAP20° |
| CNe [12] | CVPR2018 | **0.39** | 23.95 | 52.44 | 9.30 | 25.01 |
| OA-Net++ [14] | ICCV2019 | 2.47 | 39.33 | 66.93 | 16.18 | 32.40 |
| ACNe [15] | CVPR2020 | 0.41 | 33.06 | 62.91 | 18.86 | 36.04 |
| LMC-Net [17] | CVPR2021 | 0.93 | 47.50 | 73.07 | 16.82 | 34.05 |
| CL-Net [18] | ICCV2021 | 1.27 | 51.80 | 75.76 | 17.03 | 34.16 |
| MS$^2$DG-Net [19] | CVPR2022 | 2.61 | 49.13 | 76.04 | 17.84 | 36.21 |
| ConvMatch [13] | AAAI2023 | 7.49 | 57.03 | 78.74 | 21.65 | 47.04 |
| LFC [40] | ACM MM 2023 | 1.73 | 56.15 | 79.13 | 19.78 | 46.41 |
| GCT [37] | AAAI 2024 | 4.09 | 63.80 | 82.86 | 21.05 | 46.75 |
| DeMatch [41] | CVPR 2024 | 5.86 | 59.50 | 79.93 | 22.09 | 47.01 |
| LeCoT (Ours) | - | 3.74 | **65.34** | **82.71** | **22.28** | **47.05** |

Net [18], MS$^2$DG-Net [19], ConvMatch [13], LFC [40], GCT [37], and DeMatch [41]) are chosen as baselines.

**Results.** Correspondence pruning is an important foundation for subsequent computer vision tasks and we test LeCoT and baselines on outdoor and indoor scenes. As summarized in Table 1, the proposed LeCoT outperforms other methods significantly on both *Precision* (P) and the comprehensive metric $F$, showing an approximately 1 absolute percentage points over the second-best network (GCT) on outdoor scenes. Additionally, we find that the parameter quantity of MS$^2$DG-Net is most similar to our LeCoT from Table 1. Therefore, we conduct a partial comparison of typical visualization results between them in Figure 5. It is evident that LeCoT surpasses MS$^2$DG-Net in correspondence pruning to enhance network performance. Specifically, the proposed LeCoT significantly outperforms the comparison MS$^2$DG-Net in wide baselines (1*st*, 2*nd*, and 3*rd* columns), large changes in lighting (4*th* column), lack of texture information (5*th* and 8*th* columns) and repetitive structures (6*th* and 7*th* columns). In summary, the proposed LeCoT achieves optimal results in the comprehensive index $F$ under indoor and outdoor scenes for the correspondence pruning task.

**Table 3** Generalization ability test on YFCC100M, SUN3D and PhotoTourism with different feature extractors, including ORB, SuperPoint and SIFT. mAP5° is reported. Best result in each column is in bold.

| Method | Size (MB) | YFCC100M (%) | | | SUN3D (%) | | PhotoTourism (%) | |
|---|---|---|---|---|---|---|---|---|
| | | SIFT | ORB | SuperPoint | SIFT | SuperPoint | SIFT | SuperPoint |
| CNe [12] | **0.39** | 47.98 | 10.28 | 38.33 | 12.01 | 13.24 | 34.35 | 28.73 |
| OA-Net++ [14] | 2.47 | 52.18 | 14.03 | 40.25 | 14.13 | 13.52 | 42.30 | 29.85 |
| CL-Net [18] | 1.27 | 58.95 | 16.80 | 39.98 | 14.48 | 11.81 | 45.92 | 28.94 |
| MS$^2$DG-Net [19] | 2.61 | 57.68 | 14.63 | 41.68 | 15.16 | 14.11 | 46.62 | 32.94 |
| ConvMatch [13] | 7.49 | 58.95 | 15.65 | 45.07 | 15.31 | 14.23 | 48.59 | 33.41 |
| LFC [40] | 1.73 | 58.63 | 16.60 | 40.53 | 15.32 | 14.27 | 47.59 | 32.88 |
| GCT [37] | 4.09 | 62.70 | 19.40 | 37.50 | 15.73 | 11.06 | 49.05 | 27.26 |
| DeMatch [41] | 5.86 | 60.48 | 16.60 | 44.35 | 14.34 | 13.74 | 47.15 | 32.33 |
| LeCoT (Ours) | 3.74 | **64.8** | **20.45** | **45.18** | **16.35** | **14.43** | **49.85** | **34.37** |

## 4.2 Relative pose estimation

The accuracy of pose estimation can significantly reflect the performance of correspondence pruning methods. Therefore, we evaluate the proposed LeCoT and other networks on indoor and outdoor datasets preprocessed with SIFT and SuperPoint to assess their performance.

**Evaluation protocols.** We choose the weighted eight-point algorithm to estimate an essential matrix based on the predicted true correspondences, which is decomposed to a rotation vector and a translation vector. Subsequently, error metrics refer to the angular differences between the calculated rotation/translation vectors and labels. We select the mean average precision for angular errors less than or equal to 5° and less than or equal to 20° as the default metrics to evaluate the network in relative pose estimation, denoted as mAP5° and mAP20°.

**Outdoor relative pose estimation results.** The quantitative comparative experimental results of the proposed LeCoT and other state-of-the-art networks for relative pose estimation in outdoor scenes are summarized in Table 2. We choose mAP5° and mAP20° as evaluation metrics, and our LeCoT performs best on both of them. Specifically, LeCoT improves 1.54% on mAP5° than GCT (the second best network), but the number of parameters is less than half of that. Compared to MS$^2$DG-Net, which has a similar network parameter quantity to ours, LeCoT has about 16 percentage points higher on mAP5°. As shown in the first four columns of Figure 5, our LeCoT can remove masses of outliers, while MS$^2$DG-Net cannot. From this, it can be seen that the proposed LeCoT can perform well in relative pose estimation under the outdoor scenes, as it can naturally capture global context information from sparse correspondences without additional modules or data compression limitations.

**Indoor relative pose estimation results.** Additionally, we also compare these networks in indoor scenes. Due to lacking of texture information in indoor scenes, as shown in the right of Figure 5, it is more difficult to estimate relative pose. Even so, our LeCoT still performs best, as shown in Table 2. Clearly, LeCoT is 24.89% better than MS$^2$DG-Net on mAP5°. In conclusion, the proposed LeCoT obtains superior performance in the relative pose estimation task under outdoor and indoor scenes.

**Generalization ability test.** Existing two-view correspondence pruning networks often exhibit poor generalization performance to different descriptors and scenes. For instance, models trained on datasets with SIFT struggle to adapt to the datasets preprocessed with SuperPoint, or models performing well in outdoor scenes may fail in indoor scenes. However, the proposed LeCoT, leveraging a full Transformer backbone, can effortlessly capture global context information, thereby mitigating the loss of generalization ability. Specifically, all models are pre-trained on YFCC100M with SIFT and are directly evaluated on other datasets (*e.g.*, YFCC100M, SUN3D and PhotoTourism [61]) with different combinations of feature extractors (*e.g.*, SIFT, SuperPoint and ORB [62]). We use RANSAC as a post-processing step. In detail, PhotoTourism is an additional dataset, which includes a large number of photos taken by tourists during their travels, and ORB is a fast feature extractor. As shown in Table 3, we find that our LeCoT ranks first in all cases. This is because the proposed LeCoT can obtain adequate spatial and channel global context information, so that it has strong robustness and generalization abilities.

**Efficiency.** We summarize the experimental results of traditional RANSAC and several deep learning-based networks in Table 4, in which mAP5° and Average Runtime (ART, ms) are reported. Notably, these selected deep learning-based networks all achieve over 55 in mAP5° on YFCC100M with SIFT. The proposed LeCoT achieves the best performance and efficiency. Specifically, compared to RANSAC, the performance of our network is 719.60% of RANSAC, while the average runtime (ART) is only 21.05% of RANSAC. This demonstrates that the proposed LeCoT is both effective and efficient.

Table 4  Efficiency evaluation. mAP5° and the average runtime (ART, ms) of each image pair on YFCC100M with SIFT of different networks are reported. The best result on each row is in bold.

| Method | RANSAC [6] | ConvMatch [13] | LFC [40] | GCT [37] | DeMatch [41] | LeCoT |
|---|---|---|---|---|---|---|
| mAP5° (↑) | 9.08 | 57.03 | 56.28 | 63.80 | 59.50 | **65.34** |
| ART (ms↓) | 353.10 | 92.78 | 125.48 | 117.47 | 99.22 | **74.32** |

Table 5  Evaluation homography estimation on HPatches. Accuracy (ACC.) at different error thresholds is reported.

| Method | Size (MB) | HPathces (%) | | |
|---|---|---|---|---|
| | | ACC.3px | ACC.5px | ACC.10px |
| CNe [12] | **0.39** | 38.97 | 51.55 | 65.34 |
| OA-Net++ [14] | 2.47 | 39.83 | 52.76 | 62.93 |
| CL-Net [18] | 1.27 | 43.10 | 55.69 | 68.10 |
| MS$^2$DG-Net [19] | 2.61 | 41.21 | 50.17 | 62.59 |
| ConvMatch [13] | 7.49 | 43.73 | 55.78 | 66.12 |
| LFC [40] | 1.73 | 44.81 | 56.83 | 68.01 |
| GCT [37] | 4.09 | 47.58 | 57.96 | 68.25 |
| DeMatch [41] | 5.86 | 48.09 | 58.95 | 68.21 |
| LeCoT (Ours) | 3.74 | **51.93** | **60.34** | **68.59** |

### 4.3 Homography estimation

Homography estimation is a critical task in computer vision, aiming to infer the perspective transformation between images. Typically, it involves calculating a homography matrix that describes this transformation relationship by analyzing corresponding points in images. We conduct homography estimation experiments using the Direct Linear Transform (DLT) estimator on the HPatches benchmark [63].

**Dataset.** HPatches benchmark [63] comprises a total of 116 scenes, with 59 scenes exhibiting significant viewpoint changes and 57 scenes captured under varying illumination conditions. Each scene consists of 6 images, one serving as the reference image and the others as target images with ground-truth, amounting to 696 images in total, each image pair of which is detected up to 4000 keypoints with SIFT followed by a NN matcher.

**Baselines.** Eight learning-based networks (CNe [12], OA-Net++ [14], CL-Net [18], MS$^2$DG-Net [19], ConvMatch [13], LFC [40], GCT [37], and DeMatch [41]) are chosen as baselines.

**Evaluation protocols.** We follow the recommendation of SuperPoint [5], using homography error to classify the estimation results as accurate or inaccurate, and thresholds are set at 3/5/10 pixels, respectively.

**Results.** Comparison results in Table 5 indicate that the proposed LeCoT, utilizing the Transformer as the network backbone, outperforms in all scenes. Particularly, LeCoT achieves an improvement of 3.84 points over the second best network (DeMatch) on ACC.3px. This is attributed to the fact that the proposed LeCoT can obtain spatial and channel global context information among sparse correspondences, which proves advantageous for homography estimation even in challenging scenarios.

### 4.4 Visual localization

Visual localization refers to estimating 6-degree of freedom (DOF) relative pose (3-DOF for rotation and another 3-DOF for translation) from given images to a 3D model. It has many important practical applications, such as robot grasping and control, automatic navigation, augmented reality, and so on. The proposed LeCoT and other networks are integrated into the official HLoc framework [1] to accomplish visual localization.

**Table 6** Evaluation visual localization on Aachen Day-Night. We report the percentage of correctly localized queries (0.25m,2°), (0.5m,5°) and (1.0m,10°), respectively.

| Method | Size (MB) | Day | Night |
|---|---|---|---|
| | | (0.25m,2°)/(0.5m,5°)/(1.0m,10°) | |
| CNe [12] | **0.39** | 81.3/91.4/95.9 | 68.4/78.6/87.8 |
| OA-Net++ [14] | 2.47 | 82.3/91.9/96.5 | 71.4/79.6/90.8 |
| CL-Net [18] | 1.27 | 83.3/92.4/**97.0** | 71.4/80.6/**93.9** |
| MS$^2$DG-Net [19] | 2.61 | 82.8/92.1/96.8 | 70.4/82.7/**93.9** |
| ConvMatch [13] | 7.49 | 81.8/91.8/96.6 | 69.4/79.7/89.9 |
| LFC [40] | 1.73 | 82.9/92.0/96.2 | 70.8/81.3/91.3 |
| GCT [37] | 4.09 | 83.2/92.1/96.5 | 71.0/81.4/91.6 |
| DeMatch [41] | 5.86 | 83.6/**92.5**/96.7 | 71.1/81.9/92.3 |
| LeCoT (Ours) | 3.74 | **83.7/92.5/97.0** | **71.6/82.8/93.9** |

**Dataset.** We choose Aachen Day-Night benchmark [64] to test network performance. There are 922 query images (824 daytime and 98 nighttime) obtained by mobile phones and 4328 reference ones with the ancient European town style.

**Baselines.** We choose the same baselines as the homography estimation task in Section 4.3. All the networks are based on the NN strategy.

**Evaluation protocols.** Each image is extracted up to 4096 keypoints with SIFT and followed by an NN strategy. Then, a new SfM model is triangulated from the given poses and image pairs are found by image retrieval. Finally, we will perform visual localization. We choose the percentage of correctly localized queries at specific distances and orientation thresholds as evaluation matrices.

**Results.** As shown in Table 6, our LeCoT performs best in daytime and nighttime scenes with all the thresholds, especially the lowest threshold, which can demonstrate that LeCoT can work in visual localization.

### 4.5 3D reconstruction

3D reconstruction refers to the process of reconstructing the three-dimensional shape and structure of an object or scene from two-dimensional data, such as images or videos, which is widely used in fields such as virtual reality (VR), augmented reality (AR), autonomous driving, and robot navigation.

**Table 7** Comparative results of 3D reconstruction on Alamo. R+M: ratio test + mutual check.

| Method | Images | Reg. (↑) | Sparse (↑) | Dense (↑) | TL (↑) | Obs. (↑) | Reproj. (↓) |
|---|---|---|---|---|---|---|---|
| R+M | | 921 | 192714 | 3515294 | **11.22** | 2347 | **0.63**$px$ |
| OA-Net+++ [66] | 2915 | 888 | 251375 | 3164383 | 10.59 | 2997 | 0.77$px$ |
| MS$^2$DG-Net [19] | | 921 | 250328 | 3258545 | 10.46 | 2842 | 0.77$px$ |
| LeCoT | | **996** | **318021** | **3806193** | 10.76 | **2999** | 0.73$px$ |

**Dataset.** We select Alamo [65] as the test dataset, which contains 2915 images captured from the Alamo Mission in San Antonio, Texas. These images have been taken from multiple different viewpoints, aiming to capture the details of the building as well as the structure of the surrounding environment.

**Baselines.** We choose ratio test + mutual check (R+M), OA-Net+++ [66] and MS$^2$DG-Net [19] as baselines. All the networks are trained on YFCC100M with SIFT.

**Evaluation protocols.** Following OA-Net+++ [66], we choose the following evaluation metrics: 1) the number of successfully registered/aligned images (Reg.), 2) the number of points in the sparse 3D point cloud (Sparse), 3) the number of points in the dense 3D point cloud (Dense), 4) the average track length (TL), 5) the average number of observations per 3D point (Obs.), and 6) the reprojection error (Reproj.). The higher the values of metrics 1) to 5), the better the network performance, and the opposite holds for the rest.

**Results.** Following [66], we conduct 3D reconstruction experiments on the COLMAP platform [67][1]. For each image, we first extract up to 4000 feature points using SIFT descriptors, then apply various models trained on the YFCC100M dataset to filter out outliers. These filtered points are then integrated into the official COLMAP pipeline to evaluate the performance of these different models. As shown in Table 7, the proposed LeCoT achieves the best performance in terms of Reg., Sparse, Dense and Obs., and also demonstrates competitive results in the TL and Reproj., which can prove the generalizability of the proposed LeCoT.

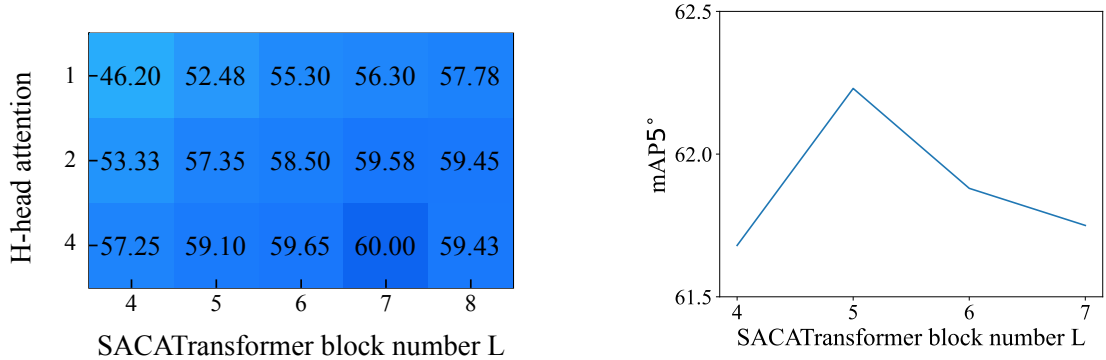

**Figure 6** (a) Relationship between the number of SACATransformer blocks and the number of attention head. mAP5°(%) is reported. (b) Relationship between mAP5°(%) and the number of SACATransformer blocks, when $H = 4$ and $po = 2$. $po=i$ represents the guiding probability set obtained at the $i\text{-}th$ SACATransformer block.

## 4.6 Ablation studies

In this section, ablation studies about LeCoT structure and its details on the outdoor dataset with SIFT are reported.

**Table 8** Evaluation on outdoor scenes with SIFT for relative pose estimation. mAP5° and mAP20° are reported. $po=i$ represents the guiding probability set obtained at the $i\text{-}th$ Transformer Block.

|  | $L = 7, H = 4$ | $L = 5, H = 4$ | | | | |
| --- | --- | --- | --- | --- | --- | --- |
|  | – | – | $po = 1$ | $po = 2$ | $po = 3$ | $po = 4$ |
| mAP5°(%) | 60.00 | 59.10 | 60.05 | **62.23** | 61.25 | 59.33 |
| mAP20°(%) | 80.61 | 80.37 | 81.02 | **81.31** | 80.61 | 80.56 |

**Relationship between $H$ and $L$.** Relationship between SACATransformer block number $L$ and attention head number $H$ has been discussed in Figure 6(a). When $H = 1$ and $L$ increases from 4 to 8, network performance gradually increases. Meanwhile, we find that when $L = 4$ and $H$ goes from 1 to 4, the network performance also gradually improves. In Figure 6 (a), the color of the squares transitions from light to dark, representing an improvement in the network performance from poor to good. We find that when $L = 7$ and $H = 4$, the network performs best, rather than simply choosing $L = 8$ and $H = 4$. Hence, we choose this combination ($L = 7$ and $H = 4$).

**How can additional guiding information help?** To obtain stronger global context information at different stages, we additionally introduce guiding information to guide network training. If the distance between the guiding information and Sorting&Pruning Block is too close, the guiding information and the pruning module information $\hat{P}_i$ may be too similar, while the distance is too far, guiding information is inaccurate. As shown in Table 8, we find that $po = 2$ is a suitable position to obtain guiding information. Observing Figure 6 (b), we can find that when $po = 2$ and $H = 4$, the model with $L = 5$ performs best.

**Is transformer structure necessary?** In the first two rows of Table 9, there is an interesting phenomenon that if we only borrow attentions from the vanilla Transformer block, network performance will rapidly decrease. This is very surprising, probably because there are no MLPs or CNNs for feature extraction. However, if the complete Transformer structure is used, the issue will disappear.

---
1) https://github.com/colmap/colmap

**Table 9** Ablation experiments about Spatial-Channel Fusion Transformer design. LeCoT (Attention) represents only using attentions instead of the fully Transformer structure. LeCoT (Vanilla) presents using the vanilla Transformer. S-MSA: Spatial Multi-Head Self-Attention; C-MSA: Channel Multi-Head Self-Attention.

|  | mAP5°(%) | mAP20°(%) |
| --- | --- | --- |
| LeCoT (Attention) | 39.88 | 60.98 |
| LeCoT (Vanilla) | 62.23 | 81.31 |
| LeCoT (C-MSA + S-MSA) | 62.30 | 81.53 |
| LeCoT (C-MSA + C-MSA) | 63.50 | 82.32 |
| LeCoT (S-MSA + S-MSA) | 63.85 | 82.38 |
| LeCoT (Ours, S-MSA + C-MSA) | **65.34** | **82.71** |

**Spatial-channel fusion transformer design.** Comparing with the third to sixth rows in Table 9, we find that utilizing Spatial Multi-Head Self-Attention (S-MSA) first and then Channel Multi-Head Self-Attention (C-MSA) is much better than utilizing C-MSA first, utilizing C-MSA twice, and utilizing S-MSA twice. Moreover, all these four combinations outperform the vanilla Transformer. This is because the proposed SACATransformer can simultaneously consider the spatial and channel similarities between correspondence features, while the vanilla Transformer cannot.

**Table 10** Ablation experiments about the prediction block. LeCoT (Simple Projection) employs the common strategy by directly applying a projection function (Conv($\cdot$)) to reduce the channel dimension from $d$ to 1 to identify inliers, while LeCoT (Ours, Prediction Block) adopts the proposed prediction block.

|  | mAP5°(%) | mAP20°(%) |
| --- | --- | --- |
| LeCoT (Simple Projection) | 64.05 | 81.98 |
| LeCoT (Ours, Prediction Block) | **65.34** | **82.71** |

**Prediction block vs. simple projection.** In Table 10, we present the performance of the proposed prediction block and a simple projection function (Conv($\cdot$)), such as in OA-Net++ [14]. It can be found that the network using the proposed prediction block outperforms significantly. This is because directly mapping features from a high dimension $d$ to 1 may result in information loss [23]. In contrast, the proposed prediction block, as shown in Figure 4, gradually reduce the dimension, thus mitigating this issue to a certain extent.

## 5 Conclusion

In this work, from a new perspective, we propose a simple yet effective two-view correspondence pruning network LeCoT, without additional module designs and data compression, which can naturally capture global context information at various stages. In particular, we propose a Spatial-Channel Fusion Transformer block as the backbone to simultaneously capture spatial and channel global context information. Additionally, we utilize the proposed prediction block to capture a probability set from the middle stage, as guiding information, which can make the proposed LeCoT to more effectively acquire global context information at different stages. Extensive experiments demonstrate that the proposed LeCoT outperforms existing state-of-the-art networks. Using Transformers as a backbone will be a new perspective in correspondence pruning, and we will look further into this in our subsequent work.

**Acknowledgements** This work was supported by the National Natural Science Foundation of China under Grant (62172226). Luanyuan Dai acknowledges the support of the China Scholarship Council program (Project ID: 202306840103).